\documentclass{article}
\usepackage{spconf,amsmath,graphicx}
\usepackage[pagebackref=true,breaklinks=true,letterpaper=true,colorlinks,bookmarks=false]{hyperref}
\usepackage{times}
\usepackage{epsfig}
\usepackage{graphicx}
\usepackage{amsmath}
\usepackage{algorithm}
\usepackage{algpseudocode}
\usepackage{amssymb}
\usepackage{enumitem}
\title{Confidence-Guided Data Augmentation for Improved Semi-supervised training}
\name{Fadoua Khmaissia and Hichem Frigui}
\address{University of Louisville}
\begin{document}
%

\maketitle
\vspace{-3cm}
\begin{abstract}

We propose a new strategy to improve the accuracy and robustness of image classification. First, we train a baseline CNN model. Then, we identify challenging regions in the feature space by identifying all misclassified samples, and correctly classified samples with low confidence values. These samples are then used to train a Variational AutoEncoder (VAE). Next, the VAE is used to generate synthetic images. Finally, the generated synthetic images are used in conjunction with the original labeled images to train a new model in a semi-supervised fashion.

Empirical results on benchmark datasets such as  STL10 and CIFAR-100 show that the synthetically generated samples can further diversify the training data, leading to improvement in image classification in comparison with the fully supervised baseline approaches using only the available data.
\end{abstract}
\vspace{-0.4cm}
\section{Introduction}
\label{sec:intro}

Deep learning models have achieved state of the art performances, especially for computer vision applications. Much of the recent successes, however, can be attributed to the existence of large, high quality, labeled datasets. In many real-world applications, collecting similar datasets is often cumbersome and time consuming.
Semi-Supervised Learning (SSL) aims to alleviate heavy labeling needs by leveraging the availability of unlabeled data to learn more robust models\cite{chapelle2009semi}. 
Data Augmentation (DA) is another solution to the problem of limited data. It aims to increase the size and variability of training datasets in order to reduce overfitting and improve the model's generalizability \cite{goodfellow2016deep}. 

Ideally, a good training set should contain enough variations within each class for the model to learn the most optimal decision boundaries.
However, when there are under-represented regions in the training feature space, especially in low data regime or in presence of low-quality inputs, the model risks learning sub-optimal decision boundaries, resulting in less accurate predictions \cite{taori2020measuring,torralba2011unbiased}. 

In this paper, we use the classification confidence of an initial model to guide synthetic data generation in order to augment the training data with samples from the under-performing regions that include the most confusing samples. We investigate the effect of generating synthetic data by a trained Variational Auto-Encoders (VAE \cite{kingma2013auto}), and the use of these as unsupervised information in a deep semi-supervised learning framework (MixMatch \cite{berthelot2019mixmatch}) avoiding thus the need to label synthetic examples. 
Our contributions can be summarized as follows:
\begin{itemize}[itemsep=0.2pt]
    \item Augment the training dataset by generating synthetic images drawn from the same distribution as those samples that are misclassified by a baseline model.
    \item Alleviate the need to label augmented data by using an semi-supervised training framework.
    \item Increase the diversity of the available training set, and thus, learn more accurate decision boundaries, without collecting additional data.
\end{itemize}

We evaluate our approach on two RGB benchmark datasets including STL10 and CIFAR100, and show that the proposed scheme improves classification performance in terms of both accuracy and robustness compared to the fully supervised baseline approaches using only the available data.

\section{Related works}

Building robust deep neural network classifiers in low data regime is still challenging despite several techniques developed to alleviate it  \cite{zhang2019dada}. Augmenting the training dataset through transformations  \cite{shorten2019survey} or generating new data by mixing existing samples and fusing their labels \cite{huang2021snapmix,yun2019cutmix} are commonly used approaches. These techniques, however, fail to cover the underrepresented regions because they only look at the direct neighborhood of input samples. Some approaches that focus on underrepresented regions include BRACE \cite{wickramanayake2021explanation} and CounterExample-based DA \cite{dreossi2018counterexample}. These approaches rely on either external repositories \cite{wickramanayake2021explanation} or physics-informed prior knowledge \cite{dreossi2018counterexample} to augment the training set. In contrast, our proposed work focuses on learning from limited data without any external data or prior knowledge.
 \begin{figure*}[ht]
\begin{center}
  \includegraphics[width=0.9\textwidth]{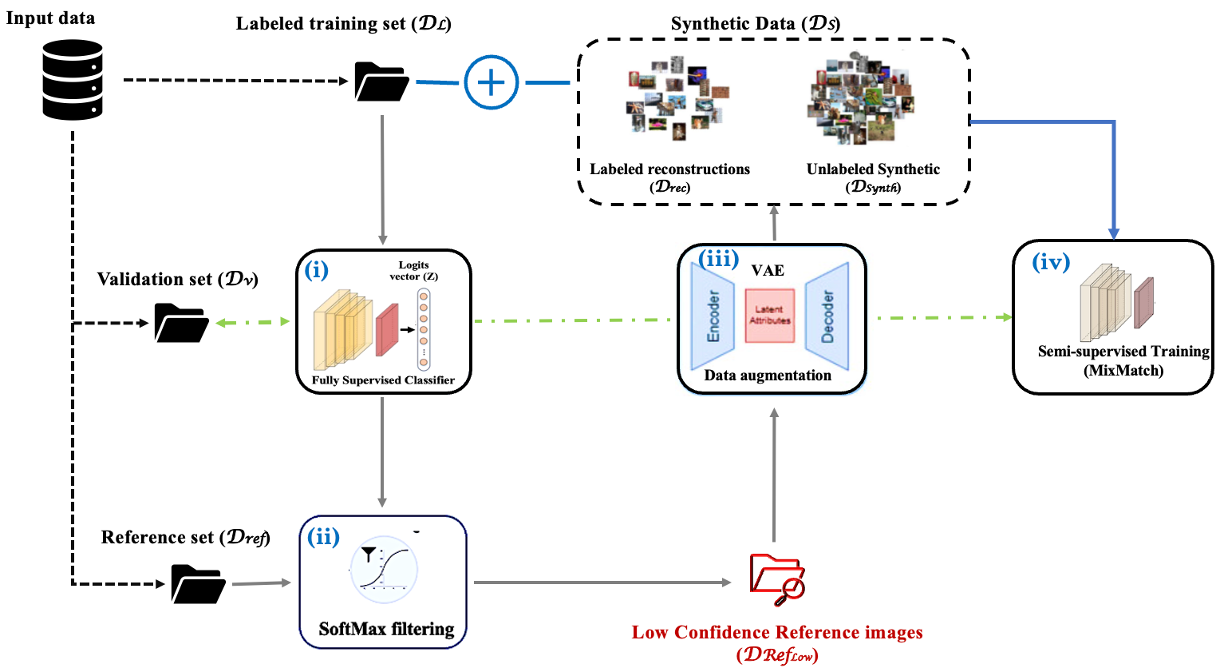}
\end{center}
  \caption{Semi-Supervised Training with Guided VAE-generated Synthetic Data: (i) Fully supervised training, (ii) Softmax filtering, (iii) Generative data augmentation, (iv) Semi-supervised training.}
\label{fig:approach}
\end{figure*}

Under ideal conditions, generative models such as GANs\cite{goodfellow2014generative} and VAEs \cite{kingma2013auto} can also be powerful tools for data augmentation. As opposed to GANS, VAEs have witnessed a limited interest to perform DA \cite{shorten2019survey,painchaud2020cardiac,selvan2020lung}.
 VAEs are trained to encode an input data point into a low-dimensional latent representation and then decode the latent representation back into the original input. By sampling from the learned latent space, VAEs can generate new synthetic data points that are similar to the training data.

Semi-supervised learning (SSL) \cite{chapelle2009semi} can help with learning in low-data regime by using both labeled and unlabeled data \cite{yang2021survey} since unlabeled data are in general easier to acquire. However, using unlabeled data can sometimes hurt model's performance if there is a distribution misalignment.

VAEs often generate blurry and fuzzy samples, especially when trained with limited data. Using VAE synthetic data as unlabeled data provides a more effective and robust approach for improving the performance of the model in semi-supervised learning, by avoiding overfitting, capturing the full data distribution, and simplifying the labeling process. Moreover, it reduces the distribution mismatch between labeled and unlabeled data that can hurt SSL performance.

\section{Proposed method}
We propose a semi-supervised data augmentation approach that generates synthetic samples sequentially and integrates them into the training process. The approach is guided by a trained fully-supervised model to generate samples from the same distribution as the least performing samples. This can be useful in scenarios with limited labeled data or when collecting additional data is not feasible.



\autoref{fig:approach} shows a diagram of the proposed approach. The training pipeline  includes four main steps: (i) Fully supervised training, (ii) Softmax confidence filtering, (iii) Generative data augmentation, (iv) Semi-supervised training. 

Formally, aside from a held out test set ($\mathcal{D}_{\mathcal{TEST}}$), we randomly split our input training dataset into three different partitions:
why not because\begin{itemize}[noitemsep]
\item $\mathcal{D}_{\mathcal{L}}=\left(\mathbf{x}_{i}, y_{i}\right)_{i=1}^{n}$ denotes the labeled training subset which contains $n$ images $x_i$ with respective labels $y_i$, where $ 1<i<n$.
\item  $\mathcal{D}_{\mathcal{V}}=\left(\mathbf{x}_{i}, y_{i}\right)_{i=1}^{n_{v}}$ denotes the validation subset of size $n_{v}$ used for model selection and hyper-parameters tuning.
\item $\mathcal{D}_{\mathcal{REF}}=\left(\mathbf{x}_{i}, y_{i}\right)_{i=1}^{n_{ref}}$, is a newly introduced subset used to select the samples which will be used to train a VAE for data augmentation.
\end{itemize}
\vspace{-0.7cm}
\subsection{Fully supervised training}
We first train and validate a fully supervised model $f_{\theta}^{FS}$ ($\theta$ denotes the model's parameter) using the labeled training set $\mathcal{D}_{\mathcal{L}}$ and the validation set $\mathcal{D}_{\mathcal{V}}$. We use WideResNet-50-2 as the fully supervised baseline model. WideResNets have achieved state of the art performances in most standard computer vision tasks, and are usually used as backbone models for deep semi-supervised models. 
\vspace{-0.4cm}
\subsection{Softmax filtering}
The model $f_{\theta}^{FS}$ is tested on $\mathcal{D}_{\mathcal{REF}}$, a separate reference subset, to identify under-performing samples. Misclassifications from $\mathcal{D}_{\mathcal{REF}}$ are expected to be similar to potential misclassifications from $\mathcal{D}_{\mathcal{TEST}}$. 
 We approximate the model's confidence score on a given prediction using $softmax$ function $S$ which converts the logits vector into a vector of probabilities, where the probabilities of each value are proportional to the relative scale of each value in the model's logits. 
We define $\mathcal{D}_{\mathcal{REF}}^{LOW}$ as follows: 
\begin{small}
\begin{equation}
    \label{eq:Dref3}
\mathcal{D}_{\mathcal{REF}}^{LOW} = \mathcal{D}^{misc}_{REF} \cup \mathcal{D}^{low}_{REF}
\end{equation}
\end{small}
where $\mathcal{D}^{misc}_{REF}$ is the set of misclassified samples by the baseline model (\autoref{eq:Dref1}), and $\mathcal{D}^{low}_{REF}$ is the set of correctly classified samples with low confidence. $\gamma$ is a user predefined confidence threshold (\autoref{eq:Dref2}). 
\begin{small}
\begin{equation}
    \label{eq:Dref1}
\mathcal{D}^{misc}_{REF}= \{ x_i \in  \mathcal{D}_{\mathcal{REF}} \mid f_{\theta}^{FS}(x_i) \neq y_y) \}_{i=1}^{n_{ref}}
\end{equation}
\end{small}
\begin{small}
\begin{equation}
    \label{eq:Dref2}
\mathcal{D}^{low}_{REF} = \{ x_i\in\mathcal{D}_{\mathcal{REF}} \mid f_{\theta}^{FS}(x_i)=y_i \And S(f_{\theta}^{FS}(x_)) \le \gamma \}_{i=1}^{n_{ref}}
\end{equation}
\end{small}


\vspace{-0.4cm}
\subsection{Data augmentation}
$\mathcal{D}_{\mathcal{REF}}^{LOW}$ is used to train a VAE in order to learn the latent distribution of the under-performing subset and generate similar synthetic samples. We pre-train the generative model on the initial training subset  $\mathcal{D}_{\mathcal{L}}$ to learn the latent representation of the target domain, and then fine-tune it on $\mathcal{D}_{\mathcal{REF}}^{LOW}$ to bias this representation more towards the under-performing samples from $\mathcal{D}_{{\mathcal{REF}}}$.
The trained VAE can generate both labeled and unlabeled data. There are two possible sets of synthetic data that we can generate:
(i) $D_{rec}^{VAE} = VAE(\mathcal{D}_{\mathcal{REF}}^{LOW})$: denotes the set of VAE reconstructions of all images in $\mathcal{D}_{\mathcal{REF}}^{LOW}$. These images are similar to their seed images and can be assigned the same labels. (ii) $D_{Synth}^{VAE}$ is a subset of $K$ randomly generated images using the VAE's decoder and cannot be labeled. 
\vspace{-.1cm}
\subsection{Semi-supervised training}
To train a semi-supervised model, we combine labeled data $\mathcal{D}_{\mathcal{L}}$ and synthetic data $D_{Synth}^{VAE}$. We use $\mathcal{D}_{\mathcal{V}}$  for both model selection and hyperparameter tuning, trying out different parameters and generative models. We use MixMatch\cite{berthelot2019mixmatch}, a technique that applies k augmentations to each unlabeled sample and generates guessed labels by averaging and sharpening the network's predictions. These guessed labels are used to pseudo-label the corresponding augmentations, and MixUp \cite{zhang2017mixup} is then used to blend the original labeled set and the pseudo-labeled set. This generates both a supervised and unsupervised loss to update the model's weights.

\section{Experimental analysis }

Our goal is to train a deep classifier on small datasets without pre-training on additional data. We evaluate our approach on RGB datasets with roughly 100-500 samples per class. We report average performances over three runs for \textbf{STL-10}\cite{coates2011analysis} and \textbf{CIFAR-100}\cite{Krizhevsky_2009_17719}. In all our experiments, $\mathcal{D}_{\mathcal{L}}$ (respectively $\mathcal{D}_{\mathcal{V}}$ and $\mathcal{D}_{\mathcal{REF}}$) constitutes 60\% (respectively 20\% and 20\%) of the input data.




The core novelty of our approach is data augmentation using generative models trained on under-performing samples that are selected based on the reference fully supervised model that is trained on the original data. For each dataset, we start by training and tuning the reference fully supervised classifier  (i.e., $f_{\theta}^{FS}$ = WideResNet-50-2) 
using $\mathcal{D}_{\mathcal{L}}$ and $\mathcal{D}_{\mathcal{V}}$. We evaluate the obtained models on $\mathcal{D}_{\mathcal{REF}}$ to identify and select the low confidence predictions: $\mathcal{D}_{\mathcal{REF}}^{LOW}$ as detailed in the previous section. 
We experiment with different confidence thresholds $\gamma$. \footnote{The value that yields the best results is the lower outlier boundary of the reference prediction scores: $\gamma = Q_1 - 1.5*IQR$ where $Q_1$ is the lower quartile and $IQR$ is the interquartile range.}

Depending on baseline performance and sample size, $\mathcal{D}_{\mathcal{REF}}^{LOW}$ may be small, making training the VAE from scratch unreliable for generating realistic synthetic images. Instead, we pre-train the VAE on $\mathcal{D}_{\mathcal{L}}$, resulting in more realistic patterns in synthetic images.



\textbf{Importance of guiding data augmentation by the under-performing samples:} 
\begin{table}[h]
\small
\begin{center}
\begin{tabular}{|c|c|c|}
\hline
Training Data &  STL-10 & CIFAR-100 \\  
\hline\hline
$\mathcal{D}_{\mathcal{L}}$ (Baseline) & 80.66$\pm$ 0.13\% & 75.75$\pm$0.09\%  \\\hline

$\mathcal{D}_{\mathcal{L}} + rand(D_{REF})^{|\mathcal{D}^{LOW}_{REF}|
}$ & 80.76 $\pm$ 0.08\% &75.93 $\pm$ 0.11\%\\
$\mathcal{D}_{\mathcal{L}}+\mathcal{D}^{LOW}_{REF}$ & \textbf{81.90 $\pm$ 0.06\%} &\textbf{76.70$\pm$ 0.05\%}\\
\hline
\end{tabular}
\end{center}
\caption{Random Augmentation vs. Guided Augmentation in a fully supervised setting: Testing accuracy across 3 runs}
\label{tab:rand_aug}
\end{table}

We investigate the importance of guiding the selected augmentations from $\mathcal{D}_{REF}$ based on the under-performing samples as identified using the reference model $f_{\theta}^{FS}$. 
We compare the performance of the fully supervised baseline using only  $\mathcal{D}^\mathcal{L}$ as labeled data versus retraining the fully supervised model using (1) The selected under-performing samples from $\mathcal{D}_{REF}$, i.e., $\mathcal{D}^{LOW}_{REF}$ (\autoref{eq:Dref3}) as additional labeled data, (2) A random subset from $\mathcal{D}_{REF}$ with same size as $\mathcal{D}^{LOW}_{REF}$ as additional labeled data.

\autoref{tab:rand_aug} displays accuracy results for the three settings on STL-10 and CIFAR-100. 
Both augmentations enhance accuracy, but using under-performing samples improves it most. This aligns with \cite{dreossi2018counterexample}'s findings that augmenting with misclassifications and low confidence correct classifications is crucial.

\begin{table}[h]
\small
\begin{center}
\begin{tabular}{|c|c|c|}
\hline
& \textbf{Fully Supervised} & \textbf{Semi-Supervised}\\
\hline
Labeled Data &  $\mathcal{D}_{\mathcal{L}} + VAE(\mathcal{D}^{Low}_{REF})$ & $\mathcal{D}_{\mathcal{L}} $\\  \hline
Unlabeled Data &  - & $VAE(\mathcal{D}^{Low}_{REF})$\\  
\hline\hline
\textbf{STL-10}& $81.33\pm0.05\%$ & \textbf{82.46$\pm$ 0.05\%}\\
\textbf{CIFAR-100}& $76.46\pm0.06\%$ & \textbf{77.12$\pm$ 0.06\%}\\
\hline
\end{tabular}
\end{center}
\caption{Using VAE reconstructions of the low confidence samples as labeled vs. as unlabeled augmentations: Testing accuracy across 3 runs}
\label{tab:VAE-aug}
\end{table}
\vspace{-0.1cm}
\textbf{Importance of using synthetic data as unsupervised knowledge:} Using VAE reconstructions as unlabeled data in addition to the original labeled data improves MixMatch's performance, as shown in \autoref{tab:VAE-aug}, compared to using them as additional labeled data to retrain a supervised classifier. This is mainly because hard labels on noisy VAE reconstructions may damage the classifier's performance. Using these reconstructions as unlabeled data enables MixMatch to learn a more robust representation of the input structure.

Since $D_{Synth}^{VAE}$ is generated from random seeds, they cannot be assigned labels, and thus, are treated as unlabeled during semi-supervised training. This can be advantageous as we can generate as many unlabeled samples as desired by sampling from different random seeds.  In our experiments we experiment with various values of $K$. 
\begin{figure}[ht]
\begin{center}
   \includegraphics[width=0.85\linewidth]{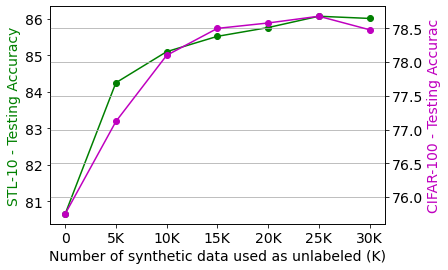}
\end{center}
   \caption{Evolution of testing accuracy on CIFAR-100 and STL-10 for different sizes of unlabeled synthetic data generated by VAE.}
\label{fig:evol-vae}
\end{figure}

Each data point in \autoref{fig:evol-vae} shows the testing accuracy  (y-axis) of a MixMatch instance that was trained on  $\mathcal{D}_{\mathcal{L}}$ as labeled and $K$ synthetic augmentations (x-axis) as unlabeled, where $K=0$ corresponds to the fully supervised baseline using the original data only. We see that, for both datasets, using VAE synthetic augmentations constantly improves the accuracy, with a peak of accuracy obtained at $K\approx25K$.


\textbf{Comparison with state of the art techniques} 
To illustrate the advantage of our ConfAugment-SST, we evaluate its performances on CIFAR100 and STL10 two slightly related data augmentation methods:
\begin{itemize}[noitemsep]
    \item \textbf{BRACE \cite{wickramanayake2021explanation}} uses concept-based explanations to augment datasets with images from an external repository, covering underrepresented regions. We adapt it to a low data regime using $\mathcal{D}_\mathcal{REF}$ as the external repository, selecting additional samples with BRACE's utility function to retrain a fully supervised model.
    
    \item \textbf{Counter Example Based DA \cite{dreossi2018counterexample}} The approach synthesizes misclassified samples to augment the training set. A domain-specific generator is introduced in the paper to create realistic samples. In our implementation, we augment the data by using all misclassifications from $\mathcal{D}_\mathcal{REF}$.
    
\end{itemize}

\begin{table}[h]
\footnotesize
\begin{center}
\begin{tabular}{|l|c|c|}
\hline
 &  STL-10 & CIFAR-100 \\  
\hline\hline
Baseline (original data) & 80.66$\pm$ 0.13\% & 75.75$\pm$0.09\%  \\
\hline
BRACE \cite{wickramanayake2021explanation} & 82.05$\pm$ 0.07\% & 77.46$\pm$ 0.05\%\\
CounterExample DA \cite{dreossi2018counterexample} & 81.01$\pm$ 0.04\% & 76.70$\pm$ 0.06\% \\
\hline
Ours (iter. 1) & 85.52 $\pm$ 0.06\% &78.50 $\pm$ 0.07\%\\
Ours (iter. 2) & \textbf{86.12 $\pm$ 0.04\%} & \textbf{79.23$\pm$ 0.05\% }  \\
\hline
\end{tabular}
\end{center}
\caption{Testing accuracy of the first two iterations of the proposed approach on STL-10 and CIFAR-100 using $K=15K$ synthetic samples. (Results averaged across 2 runs.)}
\label{tab:iter}
\end{table}

\autoref{tab:iter} shows accuracies of the baseline model, BRACE, CounterExample DA, and the first two iterations of our proposed approach on STL-10 and CIFAR-100. In the first iteration (iter. 1), we use the fully supervised Baseline to select low confidence samples from $\mathcal{D}_{Ref}$, and train the first MixMatch model. In the second iteration, we use the obtained MixMatch model as the new reference and retrain a second MixMatch model (iter. 2). We observe that our approach consistently achieves the highest accuracy for both datasets.

\autoref{tab:iter} shows also that our approach significantly improves classification performance on both datasets compared to fully supervised reference models. Further iteration yields a slight improvement, showing promise for future work. This proves that training a semi-supervised model sequentially with our proposed data augmentation technique improves upon fully supervised training with only the available data

\section{Conclusions}
We introduced a new data augmentation technique for semi-supervised training by fine-tuning a generative model based on low confidence samples from a held-out training subset. This generates both labeled and unlabeled augmentation for training a deep semi-supervised model. Experiments on CIFAR100 and STL10 show that our approach improves classification accuracy and reduces the need for additional data. Our analysis also shows that synthetic data improves the model's performance and can be extended to additional iterations.

Future work includes experimenting with different generative models and integrating the proposed components into a unified pipeline for fully supervised and semi-supervised training.

\bibliographystyle{IEEEbib}
\bibliography{strings}

\end{document}